# Low-Cost Infrared Vision Systems for Improved Safety of Emergency Vehicle Operations Under Low-Visibility Conditions

M-MAHDI NADDAF-SH[1], ANDREW LEE[2], Kin Yen[1], Eemon Amini[3], and Iman Soltani [1], (Member, IEEE)
[1]Department of Mechanical and Aerospace Engineering, University of California - Davis, Davis, CA, 95616, USA
[2]Department of Computer Science, University of California - Davis, Davis, CA, 95616, USA
[3]The Division of Research, Innovation and System Information, California Department of Transportation, Sacramento, CA, 95814, USA

Corresponding author: Iman Soltani (e-mail: isoltani@ucdavis.edu).

This work was supported by the California Department of Transportation (Caltrans).

**ABSTRACT** This study investigates the potential of infrared (IR) camera technology to enhance driver safety for emergency vehicles operating in low-visibility conditions, particularly at night and in dense fog. Such environments significantly increase the risk of collisions, especially for tow trucks and snowplows that must remain operational in challenging conditions. Conventional driver assistance systems often struggle under these conditions due to limited visibility. In contrast, IR cameras, which detect the thermal signatures of obstacles, offer a promising alternative. The evaluation combines controlled laboratory experiments, real-world field tests, and surveys of emergency vehicle operators. In addition to assessing detection performance, the study examines the feasibility of retrofitting existing Department of Transportation (DoT) fleets with cost-effective IR-based driver assistance systems. Results underscore the utility of IR technology in enhancing driver awareness and provide data-driven recommendations for scalable deployment across legacy emergency vehicle fleets.

## I. INTRODUCTION

DRIVING under low-visibility conditions presents significant safety challenges, particularly in areas with high traffic volumes and complex transportation networks [1]. At night, in poor lighting conditions, or during foggy or severe weather, such as heavy rain or snow, visibility is significantly reduced, making it difficult for drivers to detect obstacles and react to sudden hazards. This issue is especially critical for heavy emergency vehicles, such as tow trucks and snowplows, which must continue operating under these conditions to clear roadways, assist stranded motorists or those involved in accidents, and ensure transportation routes remain open [2]. Tow truck operators often need to navigate highways quickly to respond to collisions, accidental roadblocks, or breakdowns, while snowplow operators work to maintain accessibility in hazardous winter and icy conditions. Both types of emergency vehicles play a vital role in maintaining road safety, yet their ability to perform these tasks efficiently and safely is significantly hampered by limited visibility.

The reduced visibility endangers the public and the operators of these vehicles [3]. Especially in dense fog, the risk of collisions increases significantly, both for the emergency vehicles themselves and for other road users who may not see these large vehicles until it is too late [4]–[6]. Importantly, crashes occurring under fog or smoke conditions are significantly more severe than those under clear visibility, with higher rates of head-on and rear-end collisions, especially on high-speed, undivided, and unlit rural roads at night [4]. Secondary collisions involving emergency vehicles, those that occur after an initial crash, are also a significant concern, particularly in the absence of comprehensive move-over laws [7], [8]. As climate change increases the frequency of extreme weather events, these challenges are likely to become even more prevalent [9], underscoring the need for advanced technologies to improve the safety and efficiency of emergency vehicle operations in low-visibility conditions.

Most legacy emergency vehicles operated by DoTs are not equipped with modern advanced driver-assistance systems (ADAS). Even when such systems are present, their core components, namely visible spectrum cameras and radars, are primarily designed to enhance safety under standard driving conditions by detecting obstacles and issuing collision warnings [10]. However, under low visibility scenarios like dense fog, heavy snowfall, or rain, these technologies face limitations [11]. Standard cameras struggle to capture clear images in reduced visibility conditions [12]. Radar-based systems, though more robust in certain weather conditions, have limitations in detecting smaller or non-metallic objects, such as pedestrians or animals, that are commonly encountered during emergencies or in rural and unlit areas where wildlife crossings pose additional hazards [11], [13]. These limitations create critical gaps in the ability of these vehicles to navigate safely and respond effectively during emergency situations or snow-clearing operations.

IR cameras present a promising solution to the limitations faced by conventional ADAS technologies [14]–[17]. Unlike visible spectrum cameras, IR cameras detect thermal radiation emitted by objects, including other vehicles, humans and animals, regardless of the lighting conditions. Additionally, the longer wavelength of thermal radiation makes it less susceptible to scattering from small particulates in fog, rain, or snow compared to visible light, allowing IR cameras to maintain a clearer view in such conditions [18]. Combining this characteristic with the latest advances in Machine Learning (ML)-based object detection [19] makes IR technology particularly well-suited for environments where visibility is severely compromised.

The objective of this research is to evaluate the potential of IR camera technology for driver assistance in low-visibility conditions, specifically in emergency response. This study integrates controlled laboratory experiments, real-world field testing, and feedback from emergency vehicle operators, to mutually reinforce findings. Laboratory tests establish baseline capabilities of IR systems under controlled conditions, field tests confirm these capabilities in real operational environments, and surveys provide critical insights into the practical usability and operator acceptance, thus enabling comprehensive triangulation of findings. In addition to assessing the performance of IR systems in challenging conditions, the study also explores the feasibility of retrofitting existing DoT fleets with cost-effective IR-based solutions. Given that many of these vehicles are aging and lack advanced ADAS, the successful integration of low-cost off-the-shelf IR technology could improve safety. The findings of this study aim to support policymakers and transportation agencies in making data-driven decisions regarding the large-scale adoption of IR cameras to improve both driver and public safety under low visibility conditions.

The paper hereafter is structured as follows: The Related Work section reviews existing research and technologies related to driver assistance systems in low-visibility environments, highlighting gaps that this study aims to address. In the Methodology, we describe the experimental setup and procedures used to evaluate the utility of commercial IR camera systems for driver assistance. The Experimental Results section presents the findings from these evaluations, including controlled experiments, real-world trials, emergency vehicle operators' feedback, and quantitative analysis. Finally, the Conclusions and Future Work section summarizes the implications of the results, and presents suggestions for further research.

### A. RELATED WORK

Previous research has demonstrated the effect of adverse weather conditions, including rain, fog, and snow, on the performance of sensors used in ADAS [12], [20], [21]. Visible spectrum camera-based systems are particularly susceptible to challenges in conditions such as dense fog, leading to reduced contrast and visibility, which hinders the sensors' ability to accurately detect and classify objects [22], [23]. Radar systems, although more resilient in low-visibility conditions, can still experience degraded performance due to signal attenuation and scattering by precipitation, or reduced target reflectivity [24]. A similar limitation applies to LiDAR systems, whose performance can degrade due to signal scattering and absorption in rain, snow, or fog. However, LiDAR is rarely used in state-of-the-art ADAS, primarily due to its cost and integration challenges [25].

A study by Roh et al. [26] demonstrated that when rain intensity exceeded 20 mm/hour, ADAS functionality, including obstacle detection and lane-keeping, was significantly compromised. Additionally, foggy conditions have been shown to impair both the visual and LiDAR systems of autonomous and assisted driving vehicles, causing delayed obstacle recognition and hazard response [27], [28].

There has been ongoing research to explore various techniques aimed at improving the performance of these sensors in adverse conditions. One common approach involves optimizing the configuration and operation of sensors to mitigate the negative effects of rain, fog, or snow. For instance, some studies have focused on dynamically adjusting camera settings, such as exposure time, aperture (F-number), and focus, to minimize the impact of raindrops, snowflakes, or reduced contrast on image quality [29]. Another strategy to enhance sensor performance involves integrating multiple sensor modalities to compensate for the limitations of individual sensors. Sensor fusion techniques, for example, combine data from cameras, radar, and LiDAR, enabling a more robust perception system [30], [31]. In these systems, radar's resilience to precipitation and fog complements the high resolution of cameras, while LiDAR provides precise depth information, improving object detection and classification in low-visibility conditions. Deep learning algorithms have also been applied to process sensor data more effectively [32]. These methods, particularly convolutional neural networks (CNNs), can be trained to recognize and account for the effects of adverse weather, improving the accuracy of object detection and classification under challenging conditions [32]. While these techniques have shown promise in enhancing the performance of conventional sensors, the fundamental physical limitations persist [33]. No amount of post-processing or ML can fully compensate for the lack of information [29].

Research has shown that IR cameras perform well in adverse conditions, particularly when detecting objects with heat signatures, such as pedestrians, animals, and active vehicles, making them well-suited for low-visibility scenarios [34], [35]. Investigations conducted by commercial entities have also indicated that IR cameras can detect pedestrians at significantly greater distances than visible-spectrum cameras, even in complete darkness or dense fog [36]. Another focus has been on the integration of IR cameras with other sensor modalities, such as radar and LiDAR, to compensate for weaknesses in conventional systems [34], [37]. However, the use of IR cameras is not without its challenges. Environmental factors like dirt, water droplets, and ice can accumulate on the camera lens, affecting image clarity. To address this, some researchers have proposed the use of advanced data processing techniques to minimize distortions and maintain detection accuracy [38].

Despite extensive research on the potential of IR cameras, their real-world performance in driver assistance systems, especially for emergency vehicles, remains largely unexamined. Emergency vehicles face unique operational challenges, often navigating hazardous environments during rescue missions or road maintenance. These demanding conditions highlight the critical need for targeted evaluations of alternative sensor technologies. This study seeks to fill that gap.

## II. METHODOLOGY

### A. EVALUATION STRATEGY

Our evaluation strategy integrates controlled laboratory experiments, real-world field testing, and structured surveys of emergency vehicle operators to rigorously assess the effectiveness and practical usability of IR cameras under various low-visibility conditions, particularly fog. Each method addresses distinct limitations and strengths, collectively forming a comprehensive evaluation framework focused on practical, low-cost retrofitting of existing DoT fleets to enhance safety.

**Laboratory Evaluation Using Synthetic Fog:**

Controlled laboratory experiments were conducted in an enclosed tent equipped with a synthetic fog generator and provided a baseline assessment of camera performance under idealized conditions. Synthetic fog generators produce relatively smaller particulates, typically in the range of 1–10 μm, compared to natural fog, which often includes droplets ranging between 1–40 μm. Due to the smaller particulate size, synthetic fog typically results in less severe optical scattering for infrared wavelengths compared to natural fog, thus representing a somewhat optimistic scenario for IR camera performance. Therefore, any observed failure of IR cameras under these idealized conditions would definitively indicate their inadequacy for real-world applications. Conversely, strong performance under these controlled conditions suggests significant promise, yet does not guarantee equivalent outcomes in real fog scenarios with larger droplets and more complex environmental interactions. These tests also allowed for compensating for practical constraints encountered during field testing, such as seasonal, time-of-day, or logistical limitations in capturing all desired visibility scenarios.

**On-Road Testing:**

To validate laboratory-derived expectations under realistic and uncontrolled conditions, we conducted extensive field tests using a passenger test vehicle equipped with the same camera systems evaluated in the lab (Fig.1 (a)). Cameras were securely mounted side-by-side on the vehicle's roof to ensure consistent data collection under diverse, real-world visibility conditions, including clear days/nights, foggy days/nights, including extreme darkness, headlight glare from oncoming traffic, or sun glare. Field tests captured critical real-world factors not replicable in the laboratory, such as moisture condensation on camera lenses, combined fog and variable lighting conditions, and temperature fluctuations. By directly comparing field test outcomes with laboratory benchmarks, we obtained a more realistic assessment of the practical advantages and limitations of IR camera technology.

**Deployment in Emergency Vehicles and Operator Feedback:**

Given the practical constraints of simplicity and the need for cost-effective retrofitting solutions for existing DoT fleets, IR cameras were deployed in two Caltrans District 4 emergency vehicles using a straightforward and low-cost approach: integrating small in-vehicle monitors to display forward-looking IR camera feeds (Fig.1 (b,c)). Although quantitative results from laboratory and field tests suggested superior IR camera visibility under low-light and foggy conditions, the effectiveness of this simplified implementation in enhancing real-world driver performance and safety remained uncertain. Specifically, it was unclear whether operators could efficiently leverage an additional forward-looking camera feed provided on a separate monitor during actual emergency missions.

To directly address these uncertainties, structured surveys were conducted with emergency vehicle operators during routine missions. These surveys were conducted under an approved Institutional Review Board (IRB) protocol at the University of California, Davis, ensuring informed consent, voluntary participation, and anonymity of driver responses. Participants were briefed clearly about the research intent, their rights, and their ability to withdraw at any point. Due to the limited fleet size and the small pool of available operators in District 4, the participant sample was constrained. Nonetheless, operator feedback provided critical qualitative insights, revealing how well the observed laboratory and field performance benefits translated into meaningful, practical advantages for operators in real operational environments.

This evaluation framework, integrating controlled laboratory conditions, realistic field scenarios, and direct operator insights, ensured an assessment of both the technical performance and practical usability of IR camera systems, ultimately supporting informed decisions on their deployment for enhancing emergency vehicle safety under low-visibility conditions.

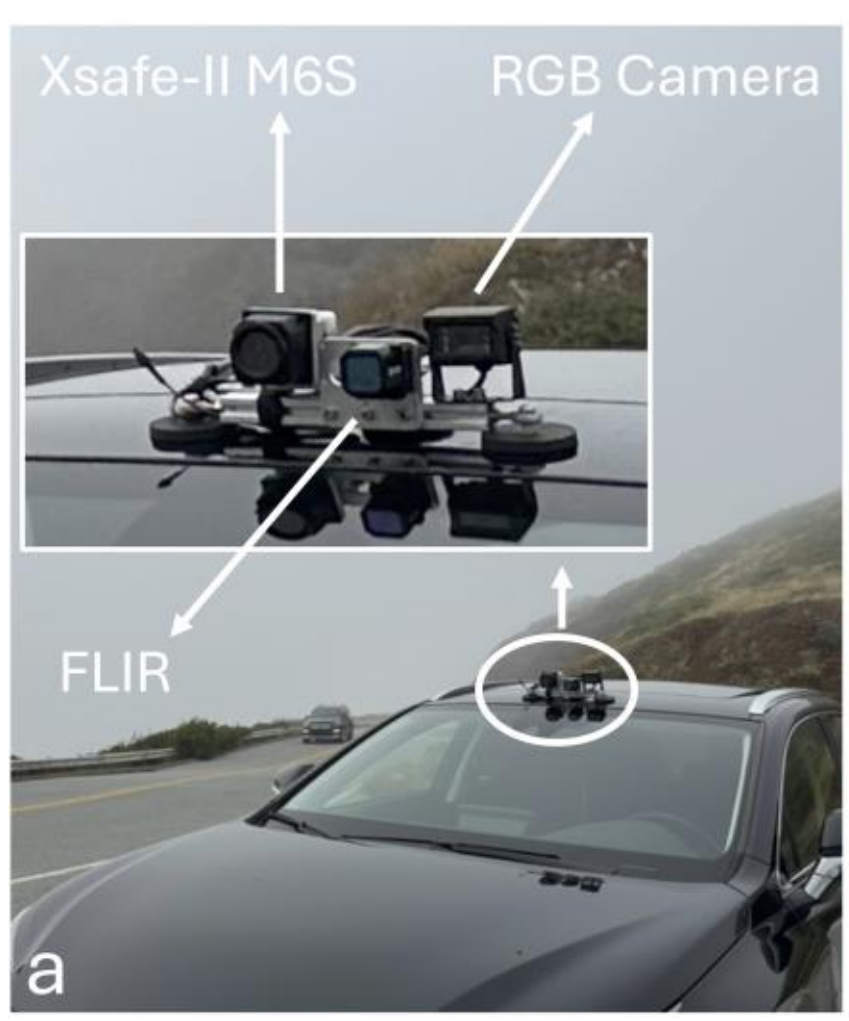

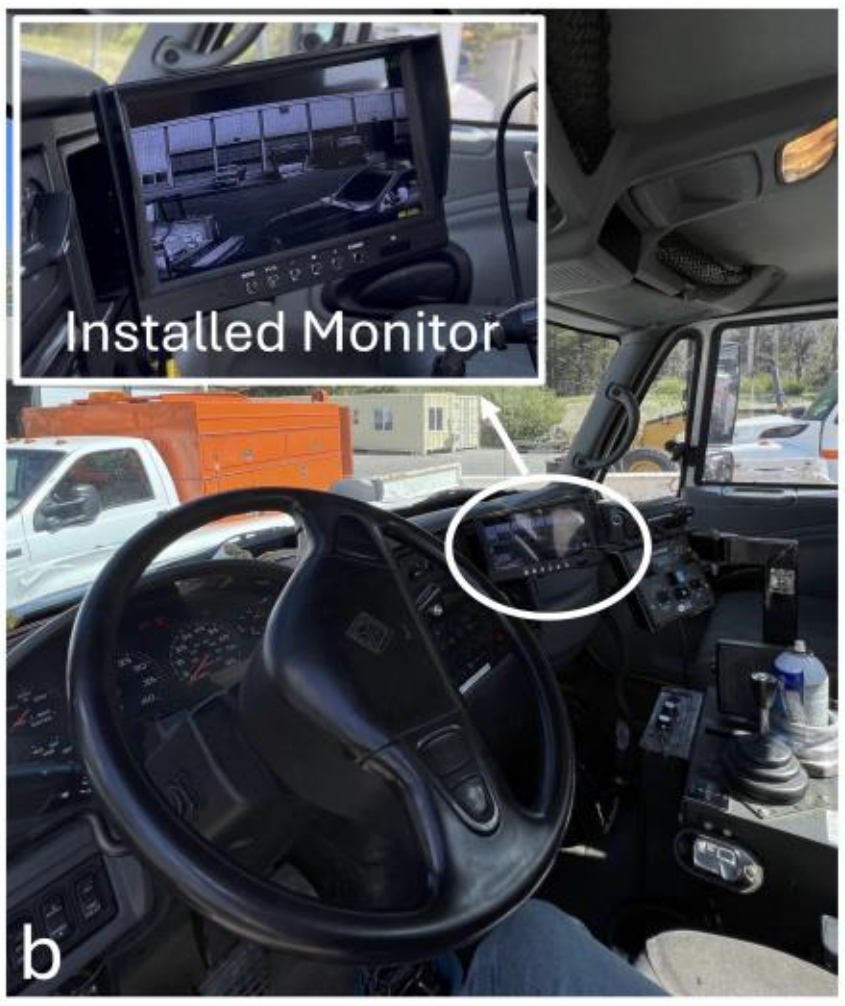

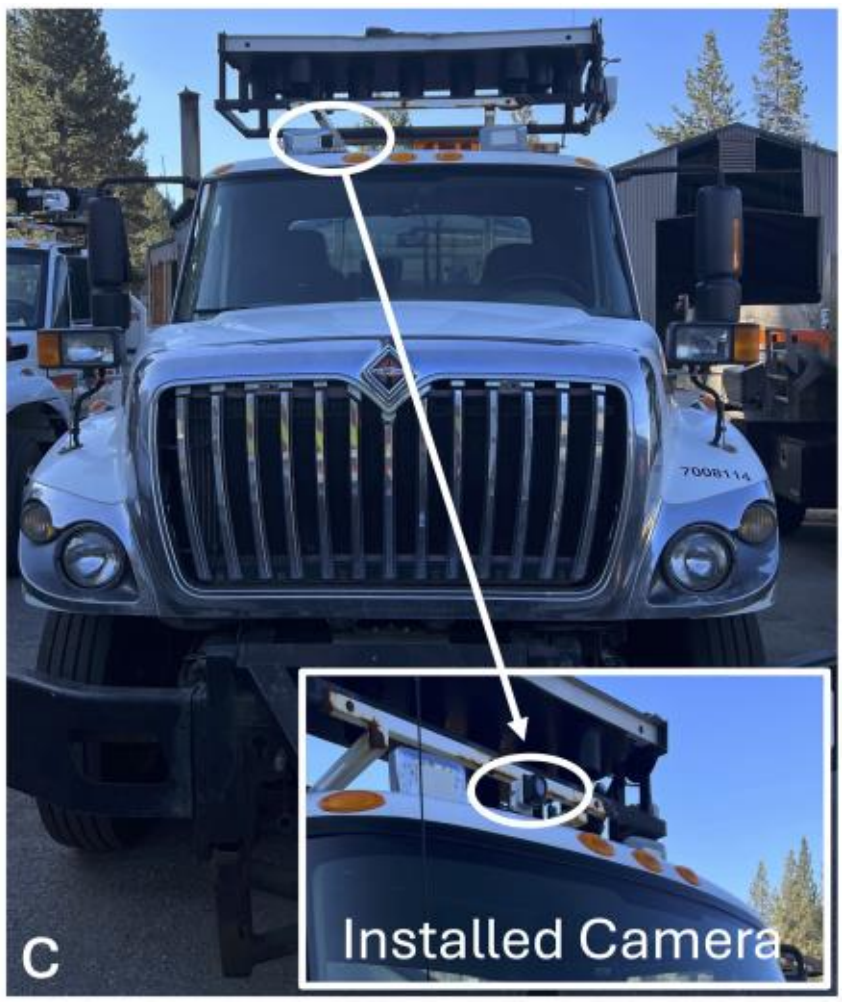

FIGURE 1. (a) Test passenger vehicle with three cameras mounted on the roof. (b) Installed monitor inside the cabin of a tow truck. (c) Installed camera on the roof of a tow truck.

### B. EQUIPMENT SELECTION

#### 1) Selection Criteria

To identify the most suitable IR cameras for our study, we conducted a comprehensive review of commercially available options. The selection process was guided by the following key criteria:

**1. Image Resolution and Quality:** The camera must provide sufficient resolution and clarity to enable the drivers as well as the ML-based object detection system to accurately identify objects in low-visibility conditions.

**2. Frame Rate:** A high frame rate is essential for minimizing presentation delay, which directly impacts the driver's total reaction time in critical situations.

**3. Connectivity:** The camera should provide a live feed output to an in-vehicle monitor and an ML processing unit for real-time object detection. The latter is necessary only if the camera system lacks embedded object detection capabilities.

**4. Cost:** Given that large-scale adoption, whether as an Original Equipment Manufacturer (OEM) solution or a retrofit upgrade, is highly dependent on cost-effectiveness, cost was another factor in the selection process.

#### 2) Selected IR Cameras

Following a comprehensive search, we selected two candidates: Teledyne from FLIR LLC [39] and Xsafe-II M6S from InfiRay Co. Ltd. [40].

FLIR Teledyne offers a resolution of 640 x 512 pixels and multiple models with various field-of-view options, ranging from 24° to 75°. The frame rate can be adjusted between 30 Hz and 60 Hz, providing real-time thermal video that can be fed directly into host systems via interfaces such as USB 2.0, Gigabit Multimedia Serial Link (GMSL), Ethernet, or Flat Panel Display Link (FPD-Link). This camera is well-suited for severe weather conditions due to its IP67-rated enclosure, which includes a heated window to prevent icing or fogging [39]. However, the FLIR ADK lacks built-in object detection capabilities, meaning that automated detection or classification of objects would require the addition of processing hardware and custom software.

Our second selection, Xsafe-II M6S, similarly provides a resolution of 640 x 512 pixels but stands out due to its built-in AI-driven object detection module. The camera is capable of detecting and classifying pedestrians and vehicles, with real-time visual bounding box overlays around detected objects on the monitor feed. This added layer of ML processing can enhance driver awareness by automatically highlighting potential pre-defined obstacles, making the system readily usable without any customization. Like the FLIR, the Xsafe-II is also designed for severe weather, featuring IP67 and IP69K ratings for durability, as well as an automatic defrost function to maintain image clarity in low temperatures [40].

### C. CUSTOM ML-BASED OBJECT DETECTION FOR TELEDYNE FLIR

#### 1) ML Hardware and Processing Needs

The Xsafe-II features an onboard object detector that is pre-programmed into the device. In contrast, the FLIR camera outputs raw video data. While this may be seen as a limitation, it offers added flexibility to integrate and evaluate the latest object detection algorithms, given the rapid pace of progress in this domain. Also certain limitations of the object detection system in the Xsafe-II can be addressed through a custom ML implementation. Notably, a delay of approximately 150 ms was observed in the Xsafe-II output, which may impact its real-time usability. In contrast, the FLIR setup offers the flexibility to optimize the object detection pipeline to reduce latency.

To process the raw video from the FLIR camera for object detection, we adopted a deep learning-based method. To ensure optimal real-time performance, we focused on minimizing delay and maximizing the frame rate in our implementation of object detection. Since this system is intended for driver assistance, where reaction time is critical, any delay in processing effectively translates to a delayed driver response, potentially increasing the risk of accidents. An analogy can be drawn to the difference in reaction times between a sober driver and one with a marginally illegal Blood Alcohol Concentration (BAC). Studies indicate that, on average, a sober individual has a premotor reaction time of around 60-90 milliseconds shorter than a person with a BAC of slightly above the legal limit [41]. This seemingly small difference can significantly impact a driver's ability to respond to sudden hazards. Applying this analogy to our system, if our object detection processing introduces an additional 100 milliseconds of delay, the driver's reaction time would be impaired to a level comparable to that of a legally intoxicated driver.

We selected NVIDIA Jetson as our ML edge computing platform due to its balance of computational power and energy efficiency for real-time processing. The FLIR camera is connected to the NVIDIA Jetson via a USB port, running YOLOv7 [42] for object detection, selected for its lightweight architecture, ease of implementation, and potential for real-time, low-latency performance [42]. To assess the trade-off between cost and performance, we evaluated different versions of YOLOv7 on two Jetson platforms: the Nvidia Jetson Nano and the more powerful and more expensive Nvidia Jetson Orin Nano.

### 2) Training Data and ML Model Optimization

For training the object detection model for the IR cameras, we used the *Teledyne FLIR Thermal Dataset* [39], which provides fully annotated thermal and visible spectrum frames captured from over-the-hood driving scenes. The dataset contains a total of 26,442 fully annotated frames with 520,000 bounding box annotations across 15 different object categories. For our application, we selected the most relevant object classes: [person, bike, car, motor, bus, train, truck, dog, deer, other vehicle], while removing all unrelated categories to streamline training and improve detection accuracy. To enhance inference speed and meet real-time application requirements, we implemented the following techniques:

**1. Model Scaling:** The Yolov7 model was optimized to align with the computational constraints of the Jetson Nano devices by adjusting its structural parameters [42]. We applied aggressive structural scaling by reducing both the number of convolutional layers and the number of channels in each layer. This resulted in a significantly shallower and narrower network with the parameter count reduced from approximately 37 million in the original YOLOv7 to 2.5 million in the scaled model. This drastic reduction lowered the computational burden and memory usage of the model, which allowed faster and more efficient inference. Importantly, these improvements were achieved while preserving the original YOLOv7 design, simply at a smaller scale. This method alone led to an approximately 2.7x increase in inference speed on the Jetson Orin Nano with minimal compromise on performance.

**TABLE 1. Overall performance of object detector algorithm for FLIR. JN: Jetson Nano, JON: Jetson Orin Nano, TRT: TensorRT. w/ and w/o indicate with and without, respectively**

| Platform | YOLOv7-tiny | YOLOv7-tiny Quantized/Scaled | YOLOv7 | YOLOv7 Quantized/Scaled |
|---|---|---|---|---|
| JN w/o TRT | 7.5 FPS | 13 FPS | - | - |
| JN w/ TRT | 13 FPS | 24 FPS | - | - |
| JON w/o TRT | - | 47 FPS | 12.5 FPS | 34 FPS |
| JON w/ TRT | - | 90 FPS | 30 FPS | 62 FPS |

**2. Quantization:** Post-training quantization was applied to reduce numerical precision in the model's weights and activations, significantly lowering computational overhead while maintaining acceptable accuracy. This optimization helped improve processing efficiency without compromising detection reliability.

**3. Inference Acceleration:** We employed NVIDIA TensorRT [43], a deep learning inference optimizer, to enhance throughput and reduce latency during model inference. TensorRT applies optimization techniques such as layer fusion and GPU-specific kernel tuning. It improved execution efficiency and significantly boosted the ML frame rate performance for the FLIR camera.

By integrating these optimizations, we substantially reduced computational complexity, enabling real-time processing at 62 Frames Per Second (FPS) on the Jetson Orin Nano despite its limited hardware resources. At this frame rate, surpassing the maximum capacity of the camera, virtually no additional delay was introduced into the camera's video throughput, ensuring minimal latency in object detection.

Table 1 summarizes the processing speed results for both hardware platforms. We began our evaluation on the Jetson Nano due to its lower cost, with the goal of identifying a viable low-budget option. Starting with the YOLOv7-tiny model, which is a smaller and less powerful variant of YOLOv7, and its quantized/scaled version, we achieved a maximum frame rate of 24 FPS, which may still be acceptable for real-time applications. However, given this upper limit, testing larger models such as YOLOv7 on the Jetson Nano was deemed impractical, as the frame rate would have dropped below usable levels. Thus, on Jetson Nano, only the tiny variant of YOLO was considered feasible for deployment in resource-constrained scenarios.

To maintain the camera's native 60 FPS throughput, we transitioned to the more powerful but costlier Jetson Orin Nano. On this platform, the quantized and scaled YOLOv7-tiny model achieved up to 90 FPS, comfortably exceeding our real-time processing target. This allowed us to evaluate more complex and powerful models like YOLOv7, which after optimization (quantization and scaling) reached a maximum of 62 FPS, sufficient to preserve the original frame rate of the input camera.

## III. EXPERIMENTAL RESULTS

### A. CONTROLLED EXPERIMENTS: SYNTHETIC FOG

To qualitatively compare performance under various levels of fog intensity, we conducted a series of controlled experiments inside a 10 × 20 ft tent equipped with an artificial fog generator. The controlled laboratory experiments using synthetic fog provided an initial baseline assessment of camera performance under idealized conditions. Synthetic fog generators produce relatively smaller particulates, typically in the range of 1–10 µm, compared to natural fog, which often includes droplets ranging between 1–40 µm. Due to the smaller particulate size, synthetic fog typically results in less severe optical scattering for infrared wavelengths compared to natural fog, thus representing a somewhat optimistic scenario for IR camera performance. Therefore, any observed failure of IR cameras under these idealized conditions would definitively indicate their inadequacy for real-world applications. Conversely, strong performance under these controlled conditions suggests significant promise, yet does not guarantee equivalent outcomes in real fog scenarios with larger droplets and more complex environmental interactions.

In our controlled experiments, both the Xsafe-II and FLIR IR camera systems consistently captured clear visuals of objects and humans across varying visibility levels—from mild to severe fog conditions, demonstrating robust performance. Additionally, the YOLOv7 object detection algorithm reliably detected objects even in dense fog conditions. In stark contrast, the RGB camera completely failed to provide usable imagery at severe fog densities, resulting in zero effective visibility. These laboratory results illustrate the substantial potential of IR camera systems to significantly enhance driver visibility and safety under severe fog. The promising outcomes observed in controlled tests justified proceeding to more comprehensive field evaluations, aimed at providing a more grounded understanding of IR camera performance under actual operational environments, detailed in the subsequent subsection. Sample frames from the synthetic fog tests highlighting these visibility differences are presented in Fig.2.

### B. FIELD EXPERIMENTS ON A PASSENGER VEHICLE

We conducted a series of field experiments to quantitatively compare the performance of the three cameras described earlier in more realistic conditions when used on a passenger vehicle. The three cameras were securely mounted on a solid aluminum bar and affixed to the roof of the test vehicle. They were positioned side by side with minimal spacing to reduce discrepancies in their fields of view (FoVs). Residual differences in the captured images, arising from displacement or misalignment, sensor size, or lens characteristics, were corrected by cropping the images to a common, aligned FoV. The evaluation was based on the detection performance of ML models, benchmarked against ground truth annotations from simultaneous video footage recorded while driving a passenger vehicle under various environmental conditions (Fig.1 (a)). RGB data was included in the comparisons to highlight the contrast between visible and infrared spectrum capabilities under varying visibility conditions. For the FLIR IR camera, we used the YOLOv7 model described in Section II-C. A separate YOLOv7 model, pre-trained on MS COCO [44], was fine-tuned on a subset of our annotated RGB footage. For the Xsafe-II system, we used the onboard object detection results provided by its embedded hardware. Detection performance across all environmental conditions is presented below.

Videos captured under various weather conditions were analyzed, with frames randomly selected and annotated as follows: 1500 frames (500 per camera) from foggy daytime conditions, 4500 frames (1500 per camera) from foggy nighttime conditions, 3000 frames (1000 per camera) from clear nighttime conditions, 1200 frames (400 per camera) from late afternoon conditions, and 4260 frames (1420 per camera) from clear day conditions. We used image timestamps to synchronize the frames from all three cameras. Table 2 presents the detection performance of all selected objects listed in II-C with respect to ground-truth human annotations. This table reports mAP@50, along with F1 score and recall calculated at a consistent confidence threshold of 0.7, which was kept fixed across all experiments. Intersection over Union (IoU) is also included to quantify the accuracy of object localization within the camera's field of view.

#### 1) Clear Day and Clear Night

This data was collected to evaluate the performance of the cameras under both typical daytime and nighttime visibility conditions. During clear daytime, vehicle speed ranged from 0 to 70 MPH, and observations were made at various times of the day, including late morning, noon, and early afternoon. Clear daytime data includes sunny and cloudy conditions. For nighttime conditions, data were collected across the same range of vehicle speeds, capturing under different lighting environments, including areas with and without streetlights, as well as scenarios with and without oncoming traffic headlights.

As shown in Table 2, under clear day conditions, all three camera systems demonstrated relatively strong performance, benefiting from high visibility. The RGB camera achieved the highest mAP@50% (0.8617), reflecting precise object localization under visible light, which is consistent with the human driver's view in ideal conditions. However, despite its slightly lower mAP (0.8103), the FLIR infrared camera outperformed the RGB camera in both F1 score (0.9395 vs. 0.8369) and recall (0.8859 vs. 0.8136), suggesting it was more consistent in detecting objects with fewer missed detections. The Xsafe-II camera showed the lowest performance among the three, with an mAP of 0.7133, an F1 score of 0.8749, and a recall of 0.7901. The IoU values were similar across all cameras, indicating that when objects were detected, the predicted bounding boxes were spatially accurate. Overall, while the RGB camera benefited from good visibility, the FLIR system still maintained robust performance, hinting at its reliability even in conditions that are already favorable for visible light cameras.

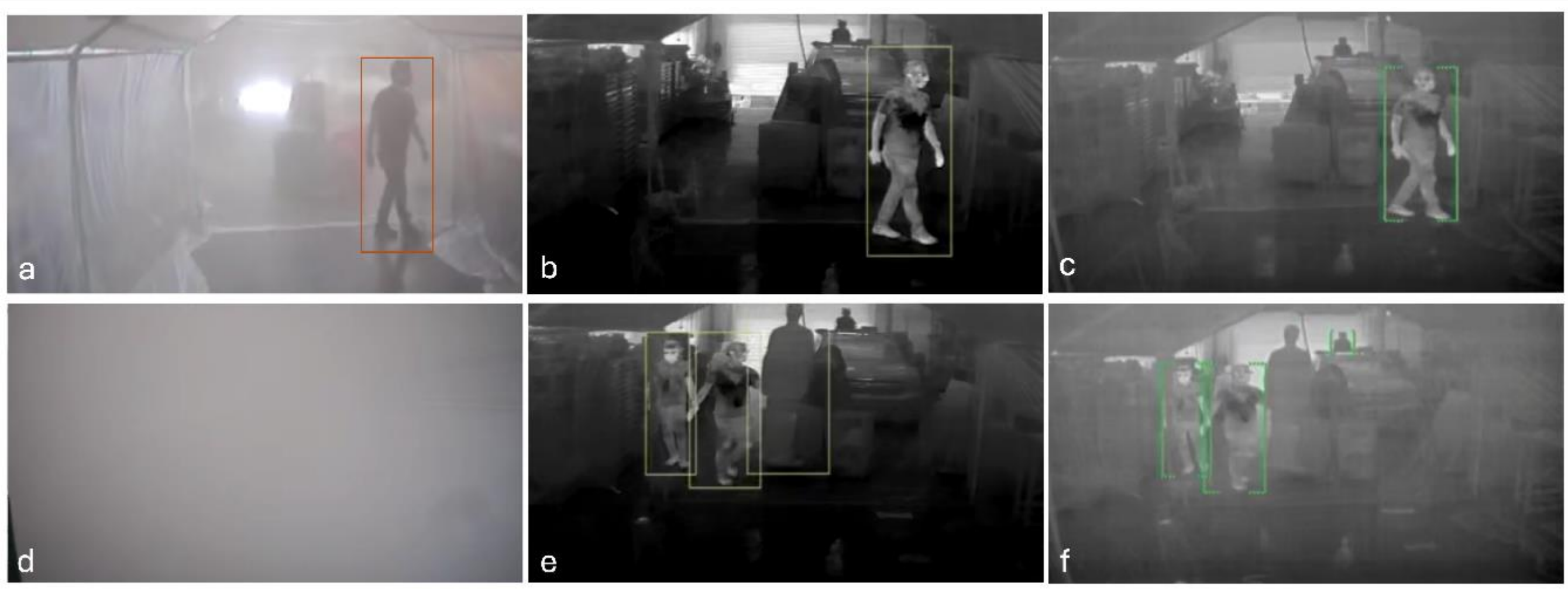

FIGURE 2. Sample frames from the synthetic fog experiment. Panels (a), (b), and (c) show RGB, Xsafe-II, and FLIR camera images, respectively, captured under light synthetic fog at the same timestamp. Panels (d), (e), and (f) present the corresponding images under dense synthetic fog, following the same camera order.

**TABLE 2. RGB Camera and IR cameras object detector performance in clear day, late afternoon, clear night, foggy day, and foggy night conditions**

| | mAP@50% | | | F1 Score | | | Recall | | | IoU | | |
|---|---|---|---|---|---|---|---|---|---|---|---|---|
| **Weather Conditions** | RGB | FLIR | Xsafe-II | RGB | FLIR | Xsafe-II | RGB | FLIR | Xsafe-II | RGB | FLIR | Xsafe-II |
| Clear Day | 0.8617 | 0.8103 | 0.7133 | 0.8369 | 0.9395 | 0.8749 | 0.8136 | 0.8859 | 0.7901 | 0.8331 | 0.8922 | 0.8912 |
| Clear Night | 0.7068 | 0.8001 | 0.7030 | 0.7044 | 0.9273 | 0.8676 | 0.7021 | 0.8861 | 0.7980 | 0.8088 | 0.8907 | 0.8882 |
| Late Afternoon | 0.7039 | 0.8924 | 0.7908 | 0.7376 | 0.9419 | 0.8759 | 0.7747 | 0.9183 | 0.8038 | 0.8166 | 0.8934 | 0.8933 |
| Foggy Day | 0.8583 | 0.7109 | 0.5419 | 0.6566 | 0.8085 | 0.7298 | 0.5317 | 0.7004 | 0.5855 | 0.8332 | 0.8912 | 0.8942 |
| Foggy Night | 0.6718 | 0.8076 | 0.6209 | 0.4737 | 0.8958 | 0.7567 | 0.3658 | 0.8330 | 0.6280 | 0.7821 | 0.8918 | 0.8920 |

In clear night conditions, the performance gap between the FLIR infrared camera and the other two systems became more pronounced. The FLIR camera maintained strong performance with an mAP@50% of 0.8001, F1 score of 0.9273, and recall of 0.8861, demonstrating its ability to reliably detect objects under low light conditions. In contrast, the RGB camera's performance dropped noticeably, with a lower mAP of 0.7068, F1 score of 0.7044, and recall of 0.7021. The Xsafe-II system achieved intermediate results, with an F1 score of 0.8676 and a recall of 0.7980. While this indicates some resilience to low-light conditions, its lower mAP (0.7030) may be attributed not only to hardware limitations but also to the efficiency and accuracy of its embedded ML model. IoU values were comparable across all cameras, suggesting similar spatial accuracy when detections occurred. Overall, these results emphasize the advantage of infrared imaging in nighttime driving scenarios. Example frames from clear day and clear night experiments are shown in Fig.3.

### 2) Late Afternoon

Part of our tests were conducted during the late afternoon hours across the same speed profiles as clear day and night. This time window was chosen intentionally due to the unique visibility challenges it presents for drivers. In particular, direct sunlight near the horizon can lead to significant glare on the windshield and the camera lens, as well as high-contrast shadows that obscure objects in the environment. These conditions differ from those typically encountered during clear midday hours, when ambient lighting is more uniform and visibility is generally high. By including late afternoon scenarios in our testing, we aimed to evaluate the performance of the system under more challenging yet commonly encountered real-world conditions.

As seen on the third row of Table 2, during late afternoon conditions, the FLIR infrared camera again demonstrated a clear advantage, achieving the highest scores across all key metrics: mAP@50% of 0.8924, F1 score of 0.9419, and recall of 0.9183. The RGB camera showed the lowest performance among the three cameras with an mAP of 0.7039, F1 score of 0.7376, and recall of 0.7747. The Xsafe-II system showed intermediate performance, with an F1 score of 0.8759, a recall of 0.8038, and an mAP of 0.7908 all lagging behind FLIR but superior to RGB. These findings underscore the late afternoon as a realistic condition where IR-based systems, like FLIR, offer superior detection reliability. Fig.4 shows examples of sun glare and high contrast shadow from the late afternoon data.

### 3) Foggy Day

This data was collected to assess the performance of camera systems in daytime foggy weather conditions. Vehicle speed varied from 0 to 55 MPH. In foggy day conditions, the performance differences between the cameras became a bit more evident. Sample images appeared blurry, and with the inten-sification of fog, detecting objects within the frame became unfeasible. Water droplets formed on the camera lenses, a consequence of the cameras being mounted externally on the vehicle and exposed to fog and humidity. The RGB camera achieved the highest mAP@50% of 0.8583. However, its F1 score (0.6566) and recall (0.5317) were substantially lower, suggesting it missed a significant number of objects due to the reduced visibility and low contrast typical of fog. In contrast, the FLIR infrared camera achieved a lower mAP (0.7109), but notably higher F1 score (0.8085) and recall (0.7004), reflecting its greater consistency in detecting objects in foggy environments. The lower mAP for FLIR, despite better recall, may reflect a higher rate of marginal detections that did not meet the IoU threshold. The Xsafe-II system showed the weakest performance, with all metrics, mAP, F1 score, and recall, falling to 0.5419, 0.7298, and 0.5855, respectively, likely due to both challenging conditions and limitations of its embedded detection model. Overall, these results suggest that the perceived advantages of IR cameras during daytime and under moderate fog conditions, consistent with our field experience (Fig.2(a, b, c)), may not be as clear-cut. In our on-road driving experiments, we primarily encountered intermediate levels of fog, where IR cameras did not show a decisive advantage over RGB. However, as discussed earlier, controlled experiments under synthetic fog confirmed that under extremely dense fog, IR cameras clearly outperform RGB (Fig.2(d, e, f)). A few examples of the foggy daytime conditions are shown in Fig.5. As illustrated in the bottom row, under moderate fog and daylight conditions, condensation on the camera lens appears to pose a more significant challenge than the fog itself.

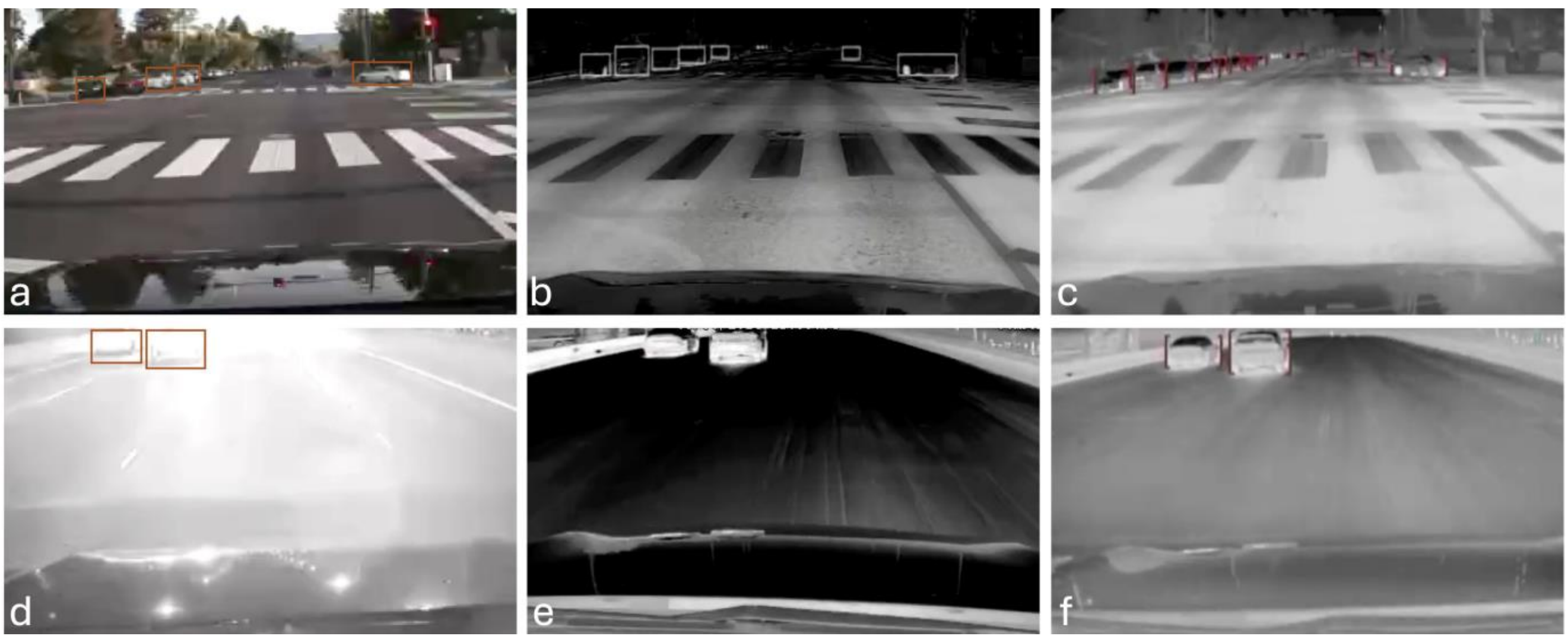

FIGURE 3. Example frames from clear day and clear night experiments. The clear day RGB, xSafe-II, and FLIR camera frame samples, at the same timestamps, are shown in (a), (b), and (c), respectively. The clear night samples in the same camera order are shown in (d), (e), and (f).

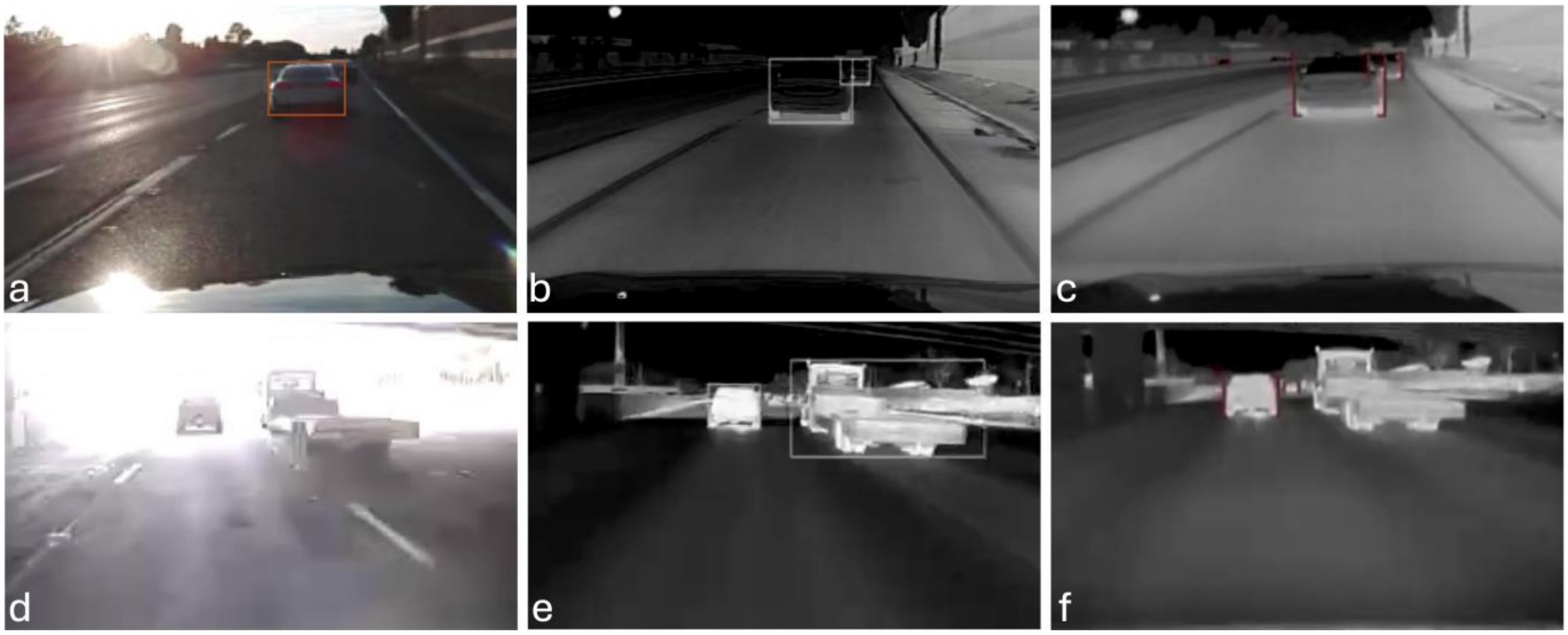

FIGURE 4. Example frames from the late afternoon dataset at two timestamps (top and bottom rows), captured using three different cameras. From left to right: RGB, Xsafe-II, and FLIR camera views, respectively.

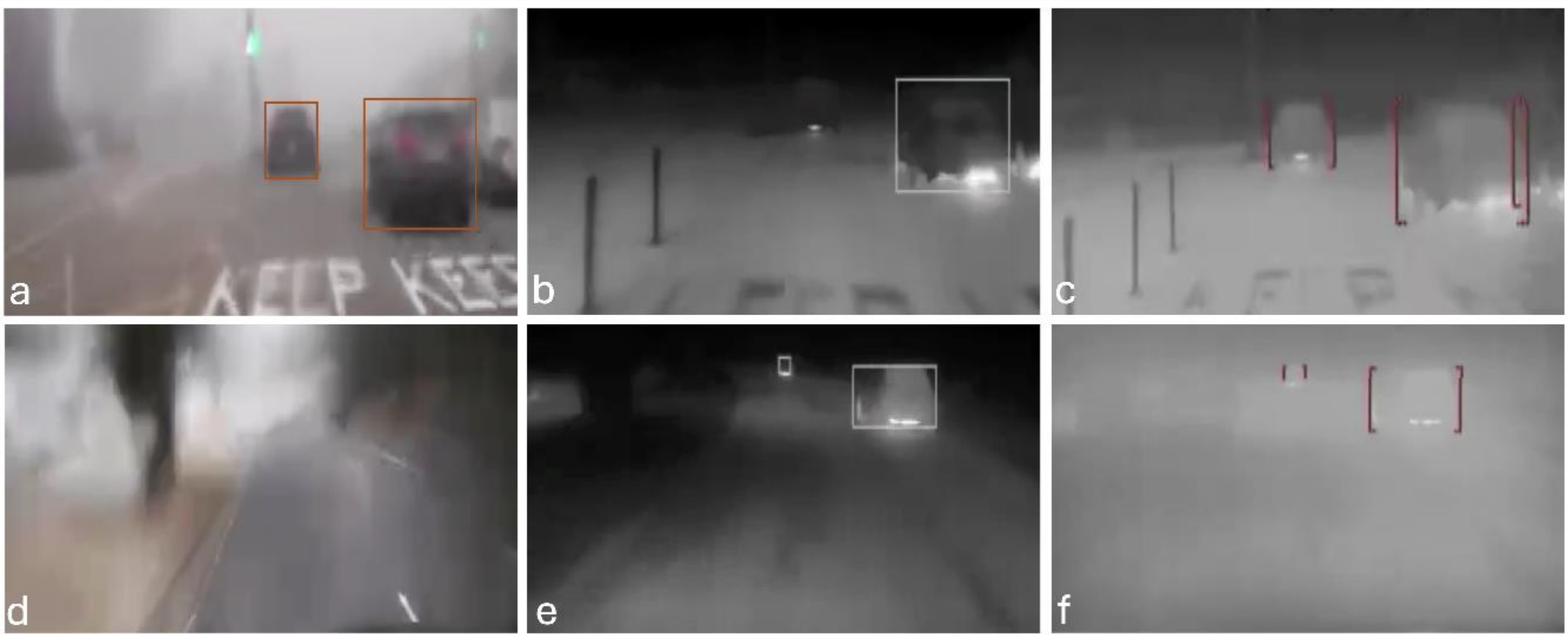

FIGURE 5. Example frames from foggy daytime conditions at two different timestamps (top and bottom rows). From left to right, the images are captured by the RGB, Xsafe-II, and FLIR cameras, respectively.

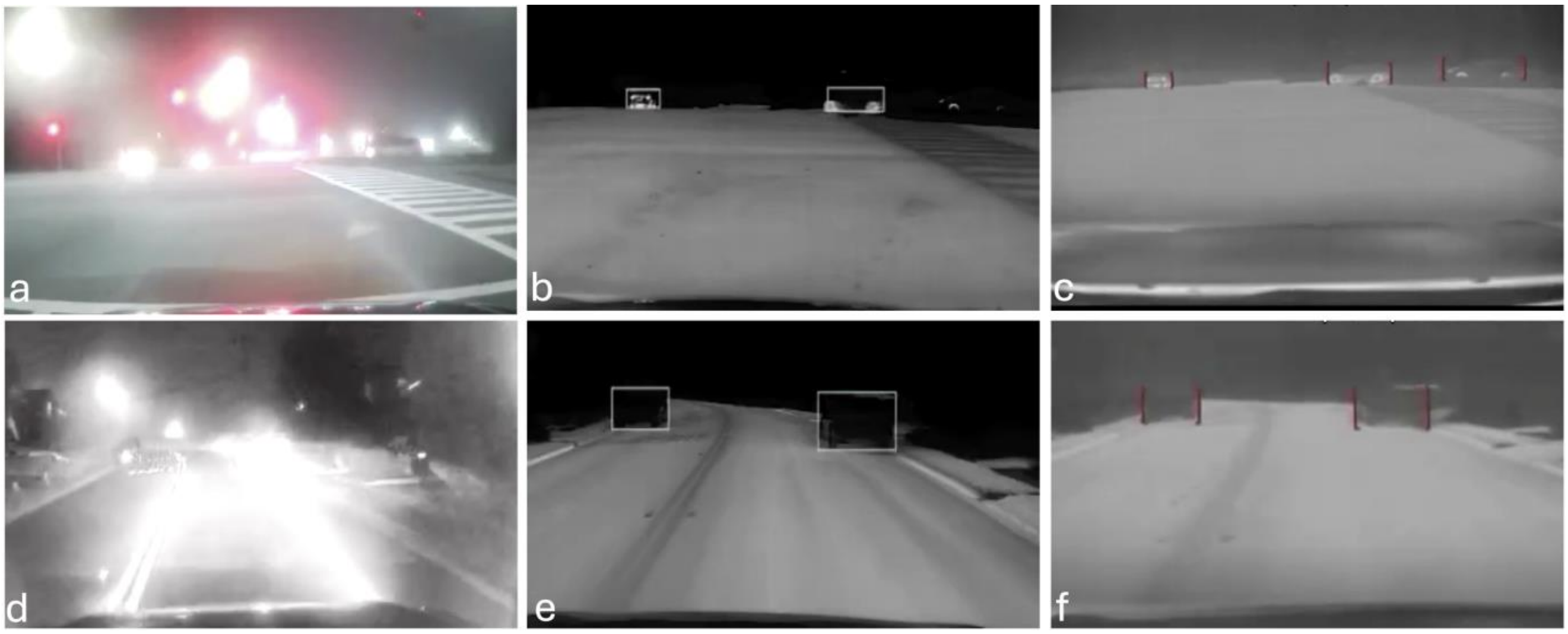

FIGURE 6. Example frames from foggy nighttime conditions at two different timestamps (top and bottom rows). From left to right, the images are captured by the RGB, Xsafe-II, and FLIR cameras, respectively.

However, this issue is not relevant to the driver's visibility inside the vehicle.

#### 4) Foggy Night

This dataset was collected to evaluate the performance of camera systems in foggy weather conditions and at night-time. Vehicle speeds were recorded, ranging from 0 to 50 MPH. Foggy night conditions posed the most challenging visibility scenario, combining low ambient light with low contrast due to fog. In this environment, the FLIR infrared camera clearly outperformed the other systems, achieving the highest F1 score (0.8958) and recall (0.8330), along with a solid mAP@50% of 0.8076. These results demonstrate FLIR's strong ability to consistently detect objects even in the most degraded visual conditions. The RGB camera, by contrast, showed a significant drop in performance, with an F1 score of just 0.4737 and recall of 0.3658, reflecting its limited capability to detect objects when both lighting and visibility are poor. The Xsafe-II camera performed better than the RGB camera but still lagged behind FLIR, with an F1 score of 0.7567 and a recall of 0.6280. Example frames from foggy nighttime conditions are shown in Fig.6.

### C. FEEDBACK OF EMERGENCY VEHICLE DRIVERS

A structured survey was conducted among eight tow truck drivers operating in the San Francisco area, California, USA, to systematically evaluate the performance and usability of the IR camera system under various conditions. Drivers rated the system across multiple criteria, including overall satisfaction, safety improvement, display quality, and system reliability, using a 1–5 rating scale (1: Poor, 5: Excellent). Table 3 summarizes the detailed quantitative and qualitative results.

**TABLE 3.** Infrared Camera System Evaluation Survey Results

| Role | Rain | Fog | Night | Overall Satisfaction | Safety Improvement | Display Quality | System Reliability | Beneficial Situations | Challenges Faced | Other Suggestions |
|---|---|---|---|---|---|---|---|---|---|---|
| Operator | 5 | 5 | 5 | 4 | 4 | 5 | 4 | Nighttime driving | None | None |
| Operator | 1 | 1 | 1 | 1 | 1 | 1 | 1 | Camera distraction | Useless in our work | Remove from wrecker |
| Operator | 3 | 3 | 3 | 3 | 2 | 3 | 2 | Low visibility/night/fog | None | None |
| Operator | 4 | 3 | 4 | 4 | 4 | 5 | 4 | Spotting pedestrians | Distracting while parking | Doesn't detect motorcycles |
| Operator | 1 | 4 | 4 | 3 | 3 | 4 | 3 | Nighttime | Poor visibility in rain | Clearer in rain |
| Operator | 4 | 5 | 5 | 5 | 3 | 4 | 4 | Nighttime | None | None |
| Operator | 4 | 4 | 4 | 2 | 2 | 3 | 3 | Inclement weather/night | Unit location | Better location |
| Operator | 4 | 4 | 3 | 4 | 3 | 3 | 2 | Low light/rain | Blurry, slow image | Brightness/color adjustment |
| **Average:** | 3.3 ± 1.5 | 3.7 ± 1.3 | 3.7 ± 1.3 | 3.3 ± 1.3 | 2.8 ± 1 | 3.5 ± 1.3 | 2.9 ± 1.1 | Nighttime, low visibility | Screen/location issues | Brightness/placement |

On average, drivers rated the system's effectiveness during fog and nighttime conditions highest, each receiving a mean rating of 3.7 out of 5, indicating a generally positive perception of its utility under these low-visibility scenarios. Performance in rain received slightly lower scores, averaging 3.3 out of 5, reflecting moderate perceived effectiveness. Overall satisfaction with the system was moderate, averaging 3.3 out of 5, while perceived safety improvements were somewhat lower, averaging 2.8 out of 5, indicating room for enhancing the system's perceived safety contributions. The calculated standard deviations (SD) highlight considerable variability among drivers' perceptions and experiences. The moderate-to-high SD values, such as those observed in ratings for rain (SD = 1.5), fog (SD = 1.3), night (SD = 1.3), and overall satisfaction (SD = 1.3), suggest notable differences in individual operator responses. This variability underscores the importance of considering personalized experiences and diverse operational contexts when deploying and evaluating the IR camera systems.

Qualitative feedback highlighted that drivers commonly found the system beneficial during nighttime and low-visibility conditions, particularly for spotting pedestrians and navigating inclement weather. However, several drivers reported challenges related to screen brightness, unit placement, and occasional image quality issues, such as blurriness and slow frame rates. Notably, the Xsafe-II system featuring embedded ML hardware was selected for these field tests primarily due to its commercial availability and presumed reliability compared to our in-house custom ML system developed for the FLIR camera. However, this embedded ML system's longer latency and lower frame rate likely contributed to driver concerns about slow imaging speed. Our custom ML implementation, by contrast, achieved nearly twice the frame rate and half the latency compared to the Xsafe-II setup, underscoring the critical importance of optimizing embedded ML processing speeds and highlighting its potential to substantially enhance driver satisfaction. One driver reported significant distraction concerns, suggesting that the system might not universally benefit all operational contexts.

Given the limited sample size and restricted operational context, these survey findings should be viewed as preliminary. Nevertheless, these insights are valuable for identifying practical challenges and informing targeted improvements for broader deployment and enhanced operator acceptance.

## IV. CONCLUSIONS AND FUTURE WORK

This study demonstrates the potential of IR camera systems as a valuable component of driver assistance technologies for emergency and maintenance vehicles. Both controlled and real-world evaluations confirm that IR cameras can reliably detect obstacles under low visibility conditions, offering a meaningful safety enhancement. The findings also underscore the feasibility of retrofitting commercial off-the-shelf IR cameras into existing vehicle fleets. Their compatibility with current fleet architectures and the relatively low integration cost make them a practical and scalable solution for transportation agencies seeking to improve safety without extensive system overhauls. Another key insight from this work relates to the rapid advancements in deep learning and object detection technologies. Rather than relying solely on embedded ML within IR cameras, utilizing higher-quality cameras with external processing pipelines may offer greater flexibility and better performance in terms of speed and latency. This decoupled approach allows for ongoing algorithmic improvements and optimization of processing throughput, further reducing latency and boosting obstacle detection performance. Future research should explore large-scale deployment of IR camera systems across diverse operational environments to further validate these results at scale and assess their impact on metrics such as collision avoidance and response times. Expanding the training dataset and retraining the object detection models with richer, more representative data could further improve detection accuracy. In parallel, efforts to refine the user experience are critical. Enhancements such as improved display quality or alternatives like auditory feedback should be explored to maximize the effectiveness and usability of the system.

## ACKNOWLEDGMENT

This research was funded by the California Department of Transportation (Caltrans).

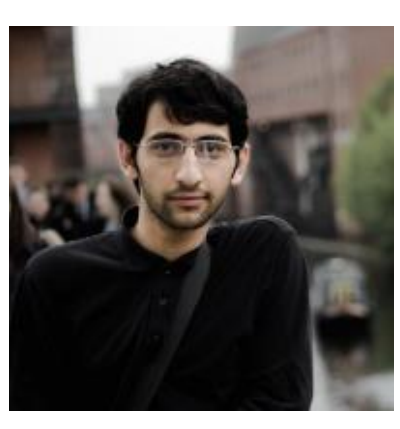

MAHDI NADDAF is a research scientist at the Laboratory for AI, Robotics, and Automation (LARA) at the University of California, Davis. His research spans the intersection of artificial intelligence, robotics, and control systems, with a focus on autonomous systems and machine learning for robotic perception and decision-making. Before joining UC Davis, he worked at Ford Greenfield Labs in Palo Alto, CA, where he led the Advanced Robotics and Automation Lab within the Core AI/ML group. He received his Bachelor's degree in Electrical Engineering from Ferdowsi University of Mashhad (Iran), his Master's in Artificial Intelligence from the University of Southampton (UK), and his Doctorate of Engineering in Electrical Engineering from Lamar University (USA). He has authored several publications in peer-reviewed IEEE journals and conferences, with contributions recognized in the areas of autonomous robotics and intelligent manufacturing.

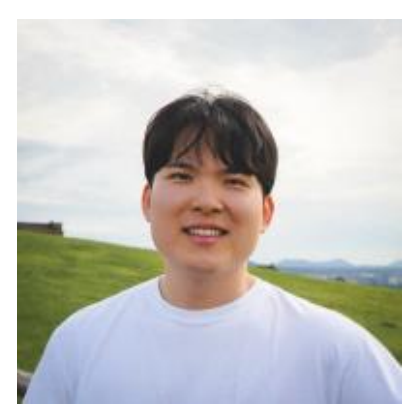

ANDREW LEE received the B.S. degree in mechanical engineering from Hanyang University, Seoul, South Korea. He is currently pursuing the Ph.D. degree in computer science at the University of California, Davis, Davis, CA, USA. He is a Graduate Student Researcher with the Laboratory for AI, Robotics and Automation (LARA) at UC Davis. His research interests include robot learning, human-inspired sensorimotor integration, active vision, and tactile sensing for robotic systems.

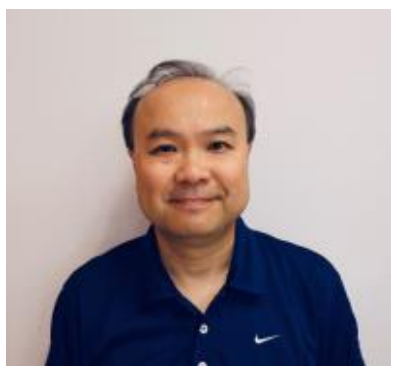

KIN YEN earned his bachelor's degree in mechanical engineering from the California State University at Fresno, California, US in 1994. He completed his master's degree in mechanical engineering from the University of California-Davis in 2002. He is a senior development engineer at the Advanced Highway Maintenance and Construction Technology Research Center, University of California, Davis since 2002. He has over 23 years of experience applying advanced automation, robotics, remote sensing, surveying, and related technologies for transportation system operations and maintenance. He has extensive project lead and participation experience with Caltrans projects including static and mobile terrestrial laser scanning systems (MTLS), bathymetric survey, Uncrewed Aerial System mapping, geographic information systems, driver assistance systems, and vision-based systems for automated site monitoring.

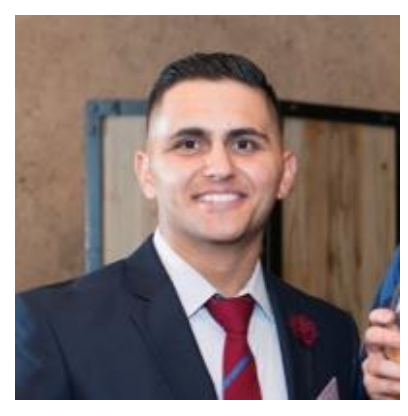

EEMON AMINI received the B.S. degree in Civil Engineering from California State University, Sacramento, in 2013, and the MBA degree with a focus on Management from Strayer University in 2015. He is currently a Research Project Manager with Caltrans, Sacramento, CA, USA, where he manages research projects focusing on advanced driver assistance systems and infrastructure safety technologies. His professional interests include project management, product innovation, and continuous improvement methodologies.

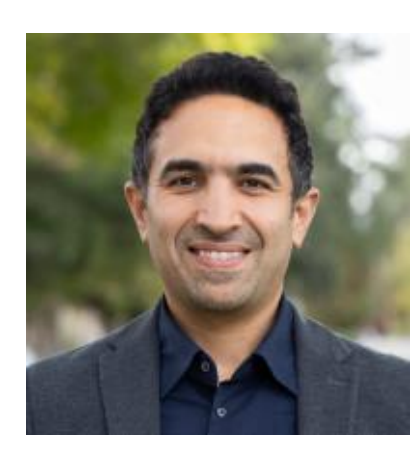

IMAN SOLTANI is an assistant professor of Mechanical and Aerospace Engineering and a faculty member in graduate groups of the Departments of Computer Science and Electrical and Computer Engineering at the University of California, Davis. His research spans the interface of artificial intelligence, instrumentation, controls, and design, with a focus on machine learning and robotic systems. Before joining UC Davis, he worked at the Ford Greenfield Labs in Palo Alto, CA, where he founded and led the Advanced Automation Laboratory. He earned his bachelor's, master's, and PhD in mechanical engineering from Tehran Polytechnic (Iran), the University of Ottawa (Canada), and the Massachusetts Institute of Technology (MIT), respectively. He holds more than 18 patents and has authored over 40 journal and conference publications on topics ranging from medical imaging to autonomous driving, high-speed nanorobotics, dexterous bimanual robotics, machinery health monitoring, and precision positioning systems. His research has been featured in prominent outlets such as The Boston Globe, Elsevier Materials Today, ScienceDaily, and MIT News. Among his numerous awards are the MIT Carl G. Sontheimer Award and National Instruments' Engineering Impact Award.

• • •